\documentclass[conference]{IEEEtran}
\IEEEoverridecommandlockouts

\usepackage{cite}
\usepackage{amsmath,amssymb,amsfonts}
\usepackage{algorithmic}
\usepackage{graphicx}
\usepackage{textcomp}
\usepackage{caption}
\usepackage{multirow}
\usepackage{subcaption}
\usepackage{booktabs}
\usepackage{xcolor}
\usepackage{comment}
\usepackage{placeins}
\usepackage{makecell}   
\usepackage{array}      
\usepackage{adjustbox}
\usepackage{arydshln}
\usepackage[utf8]{inputenc}\usepackage{float}

\def\BibTeX{{\rm B\kern-.05em{\sc i\kern-.025em b}\kern-.08em
    T\kern-.1667em\lower.7ex\hbox{E}\kern-.125emX}}
\begin{document}

\title{Multi-Crop Leaf Disease Recognition:
A Unified Benchmark and Cross-Region Study \\

}
\author{\IEEEauthorblockN{1\textsuperscript{st} Rosemary Nalwanga}
\IEEEauthorblockA{\textit{Data Management and Biometrics} \\
\textit{University of Twente}\\
Enschede, Netherlands \\
rosemary.nalwanga@utwente.nl}
\and
\IEEEauthorblockN{2\textsuperscript{nd} Sebastian Bunda}
\IEEEauthorblockA{\textit{Data Management and Biometrics} \\
\textit{University of Twente}\\
Enschede, Netherlands \\
s.t.bunda@utwente.nl}
\and
\IEEEauthorblockN{3\textsuperscript{rd} Godliver Owomugisha}
\IEEEauthorblockA{\textit{Department of Computer Engineering} \\
\textit{Busitema University}\\
Tororo, Uganda \\
ogodliver.eng@busitema.ac.ug}
\and
\IEEEauthorblockN{4\textsuperscript{th} Luuk Spreeuwers}
\IEEEauthorblockA{\textit{Data Management and Biometrics} \\
\textit{University of Twente}\\
Enschede, Netherlands \\
l.j.spreeuwers@utwente.nl}
\and
\IEEEauthorblockN{5\textsuperscript{th} Estefania Talavera Martinez}
\IEEEauthorblockA{\textit{Data Management and Biometrics} \\
\textit{University of Twente}\\
Enschede, Netherlands \\
e.talaveramartinez@utwente.nl}
}
\maketitle

\begin{abstract}
Deep learning models for crop leaf disease recognition routinely report near-perfect accuracy yet are typically trained and evaluated on a single dataset collected under controlled laboratory conditions, leaving their behavior under realistic cross-region domain shift poorly understood. We introduce MLD (Multi-crop Leaf Disease) dataset, a unified multi-region benchmark that combines six public crop-disease datasets from the USA, Asia, and Africa into a shared hierarchical taxonomy spanning 18 crops, 56 crop-disease classes (including one healthy class per crop) making 167,427 images. We define standardized single-source and pooled multi-source evaluation protocols that explicitly probe cross-region generalization. We also investigate whether exploiting the inherent crop-to-disease dependency via a hierarchical formulation (HiLeaD) that conditions disease prediction on the predicted crop improves recognition under cross-region shift. 
Under the HiLeaD, the model trained on PlantVillage achieves 99.07\% in-domain disease F1 but collapses to 12.88\% when tested on PlantDoc, exposing a severe cross-region domain gap. The model trained on the pooled MLD dataset partly recovers cross-region disease F1 from 12.88\% to 39.64\% on PlantDoc (HiLeaD), achieving a 26.76 percentage point improvement. The hierarchical formulation provides a consistent additional gain, ranging from 1.71 to 4.88 percentage points in disease F1 over the flat baseline under the MLD dataset indicating that progress in this area is currently limited more by data coverage and diversity than by model design.
\end{abstract}


\section{Introduction}
Crop diseases are a major threat to global food security and to smallholder livelihoods in particular. Globally, especially in sub-Saharan Africa and Latin America, smallholder farmers who typically grow several staple crops on plots of less than one hectare depend on these crops for both food and income \cite{giller2021small}. Disease outbreaks can wipe out a substantial fraction of yield before farmers can intervene \cite{hlatshwayo2023determinants}.  Many smallholder farmers cultivate multiple crops simultaneously, so a practical decision-support tool must be able to diagnose diseases across several crops rather than a single species \cite{islam2024determinants}, \cite{basantaray2024crop}. Leaf imaging is a natural signal for such tools where chlorosis, necrotic lesions, and other foliar symptoms are visible early and reliably correlate with the underlying pathology \cite{gao2025symptom}, \cite{shalini2025machine}. 

\begin{figure}[ht!]
    \centering
    \begin{subfigure}{0.15\textwidth}
        \centering
        \includegraphics[width=\linewidth]{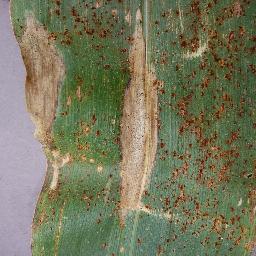}
        \caption{(PV) Corn northern leaf blight}
        \label{fig:apple11}
    \end{subfigure}
    \hfill
    \begin{subfigure}{0.15\textwidth}
        \centering
        \includegraphics[width=\linewidth]{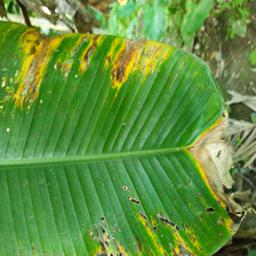}
        \caption{(PM) Banana yellow sigatoka}
        \label{fig:grape42}
    \end{subfigure}
    \hfill
    \begin{subfigure}{0.15\textwidth}
        \centering
        \includegraphics[width=\linewidth]{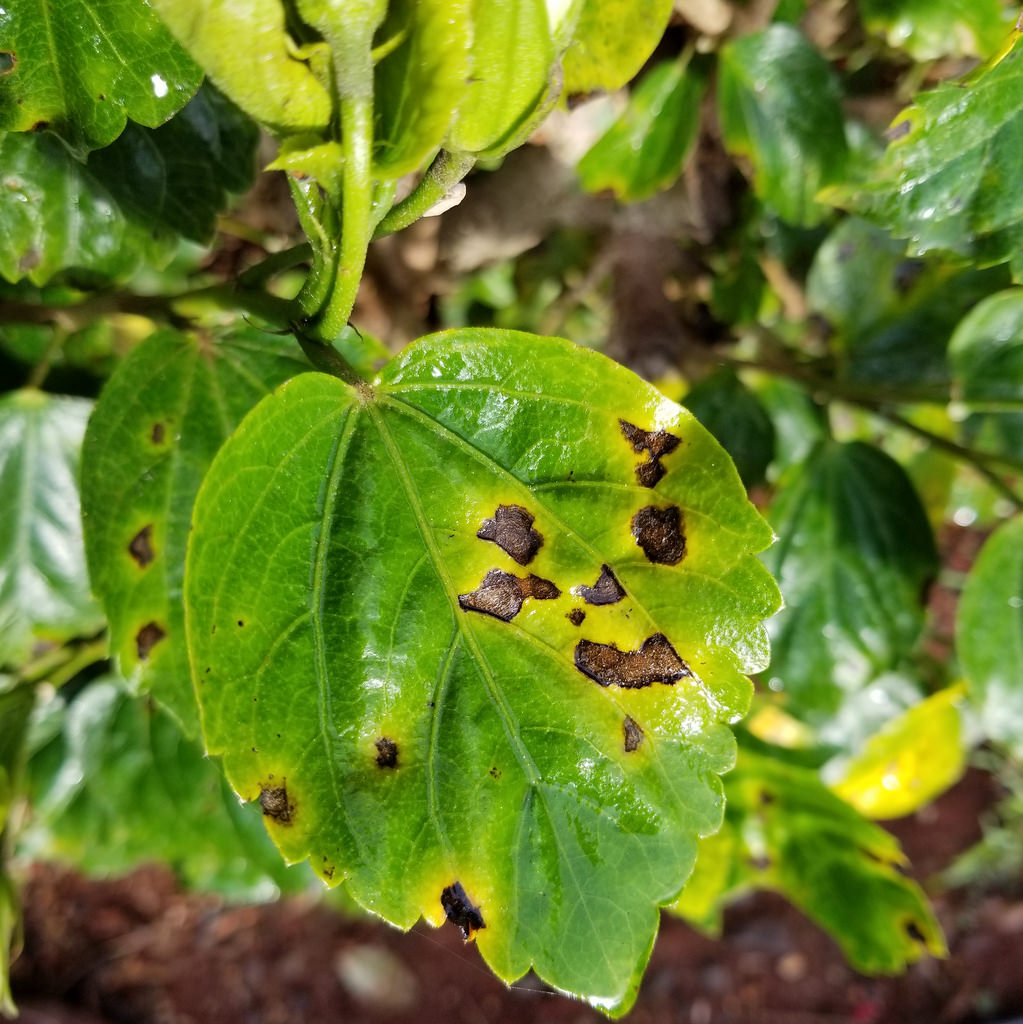 }
        \caption{(PD) Tomato bacterial spot}
        \label{fig:spot14}
    \end{subfigure}

     \vspace{0.2em} 
     
    \begin{subfigure}{0.15\textwidth}
        \centering
        \includegraphics[width=\linewidth]{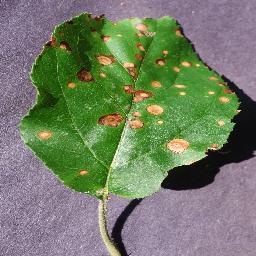}
        \caption{(PV) Apple black rot}
        \label{fig:corn46}
    \end{subfigure}
    \hfill
    \begin{subfigure}{0.15\textwidth}
        \centering
        \includegraphics[width=\linewidth]{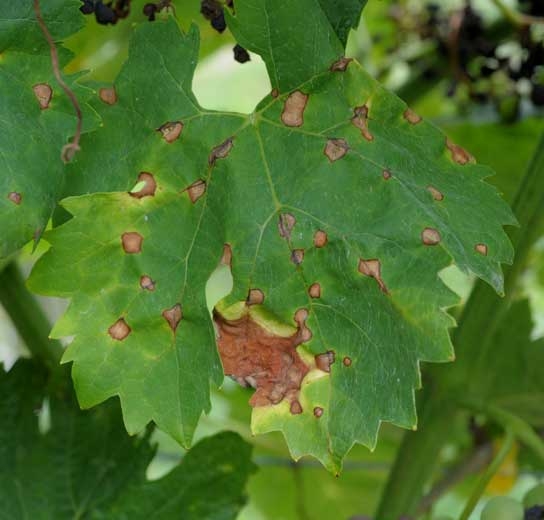}
        \caption{(PD) Grape black rot}
        \label{fig:banana17}
    \end{subfigure}
    \hfill
    \begin{subfigure}{0.15\textwidth}
        \centering
        \includegraphics[width=\linewidth]{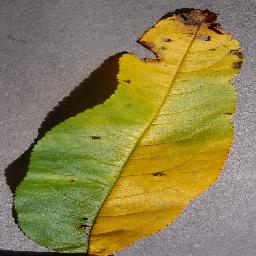}
        \caption{(PV) Peach bacterial}
        \label{fig:peach18}
    \end{subfigure}

    \caption{Image samples for diseased leaves from the different datasets where PV = PlantVillage\cite{mohanty2016using}, PD = PlantDoc\cite{singh2020plantdoc} and PM = PSFD-Musa\cite{medhi2022psfd}.}
    \label{fig:all_crop_images1}
\end{figure}

Recent advances in computer vision and machine learning have enabled accurate crop disease detection using Convolutional Neural Networks (CNNs), Vision transformers, and hybrid models, with classification accuracies ranging from 84\% to 99.8\% across real-world datasets \cite{pathak2025cnns}. However, their practical usability is severely limited by poor generalization across crops, datasets, and regions. Most approaches are single crop focused under controlled environments, lacking real-world applicability for smallholder farmers with diverse fields. Ideally, farmers need a single pipeline model that can diagnose diseases across multiple crops grown on their fields. Models trained on one crop rarely generalize well to other crops, due to differing physiological traits and overlapping symptoms among diseases as seen with samples in Fig. \ref{fig:all_crop_images1}. Several benchmark crop leaf datasets, the most popular being PlantVillage \cite{mohanty2016using}, have demonstrated high potential with CNNs emerging as the dominant architecture across most studies. However, some diseases share patterns across crops \cite{dilliwar2025plant} which complicates disease recognition across crops with visually similar symptoms.

Deep learning models trained on single-source images usually report high accuracies on standard crop-disease datasets. The reported numbers, however, are misleading as estimates of field performance. The vast majority of crop-disease models are trained and evaluated on a single dataset collected under controlled lab conditions with uniform backgrounds and consistent acquisition protocols. When the same models are applied to images from a different region, captured with different cameras under field conditions, the performance drops sharply \cite{moupojou2024segment}.

Additionally, when the label space is typically treated as a flat list of categories, it ignores the natural crop-to-disease relationships and the fact that multiple datasets may use overlapping but non-identical taxonomies. Crops and diseases are naturally hierarchical, with each disease associated with a specific crop rather than multiple crops.

The gap our work addresses is the absence of a unified multi-region benchmark and a systematic investigation of the crop-to-disease dependency structure under cross-region domain shift.

There is currently no multi-region, unified multi-crop benchmark that (i) combines heterogeneous public datasets into a shared crop-disease taxonomy and (ii) examines whether exploiting the crop-to-disease dependency improves recognition under such shift.

In this work, we introduce MLD (Multi-crop Leaf Disease) dataset, a unified multi-region benchmark for multi-crop leaf disease recognition. We combine six public datasets spanning the USA, India, Ghana, Uganda, and Tanzania into a shared crop–disease taxonomy covering eighteen crops, and  define standardized single-source and pooled multi-source evaluation protocols that probe cross-region domain shift. Within this benchmark, we compare flat and hierarchical classification formulations and analyze how data diversity affects generalization.

The main contributions of this work are:
\begin{enumerate}
    \item A unified multi-region benchmark, MLD dataset that combines six public crop-disease datasets from three continents into a shared hierarchical label space across eighteen crops.
    \item A standardized single-source and pooled multi-source cross-region evaluation protocol that measures performance under domain shift.
    \item An empirical analysis comparing flat and hierarchical classification formulations, and a study of how data diversity affects generalization across regions.
\end{enumerate}

\section{Existing works}
Recent multi-crop disease detection research has demonstrated significant advances in machine learning techniques for plant disease recognition. Researchers have developed sophisticated models achieving remarkable accuracy. Sajjad Saleem et al.\cite{saleem2024comparison} introduced an AgirLeafNet, achieving 100\% accuracy for potato, 92\% for tomato, and 99.8\% for mango leaf disease detection. Zubair Saeed et al. \cite{saeed2021multi} developed CNN-based approaches for rice and corn, reaching accuracies of 99.10\% and 97.81\%, respectively. A study in \cite{hamid2025enhancing} evaluated multi-model approaches across potatoes, tomatoes, grapes, apples, and peaches, with the VGG16 model achieving 99\% accuracy. While several studies address multiple crops, they often develop specialized models for individual crops first. A study in \cite{sharma2023dlmc} tested a deep learning multi-class classification model on individual datasets for citrus (93.56\% accuracy), cucumber (92.34\%), grapes (99.50\%), and tomato (96.56\%), treating each as a separate classification problem before combining them. A comprehensive framework, ‘MaizeNet’ for maize crop disease detection was developed \cite{kundu2022disease}, achieving 98.50\% accuracy. 

Several other studies have attained high accuracies on crop disease detection, yet existing work has either mostly focused on single-crop at at time or single-region scenarios \cite{nalwanga2025multi}. As a result, models that achieve close to $100\%$ accuracy on in-domain benchmarks may fail when tested on images from other different datasets, regions or field conditions \cite{pathak2025cnns}. 
Despite recent advances in crop leaf disease detection, several gaps persist.
\paragraph{Single-Crop Focus}
Most existing models are designed for single-crop disease detection and these fail when applied to other crops. 

\paragraph{Limited generalization across domains}
State-of-the-art models often achieve and report near-perfect high accuracy with in-domain datasets, but their performance degrades sharply when tested on data from different regions, acquisition devices, or environmental conditions.

\paragraph{Fragmented datasets and insufficient data diversity}
Most prior work evaluates models on single-source datasets, with no exploration of how dataset aggregation from diverse regions can improve generalization and robustness.

\paragraph{Lack of interpretable, modular models}
Few existing approaches offer architectures that mirror agronomic reasoning (i.e., first identify the crop, then the disease), which would improve interpretability and enable modular extension to new crops or diseases without full retraining. \\
The discussed works in this section mostly present performance of models trained and tested on single-source datasets with uniform acquisition conditions, leaving cross-region behavior unmeasured. Secondly, the crop-disease label space is treated as flat, ignoring the natural hierarchy. These gaps motivate our MLD benchmark work in this paper.

\section{MLD: Multi-crop Leaf Disease dataset}
\label{data}
Crops across various datasets of PlantVillage \cite{mohanty2016using}, PlantDoc \cite{singh2020plantdoc}, CCMT \cite{mensah2023ccmt}, Harvard Dataverse datasets of Makerere AI Lab (MAK) \cite{babirye_makerere_maize_2022}  and Nelson Mandela African Institute of Tanzania (NMAIST) \cite{mduma_nmaist_bananas_2022}, and PSFD-Musa \cite{medhi2022psfd} have been used in the benchmark. These datasets originate from the USA, Asia, and Africa. Table~\ref{tab:datasets} summarizes the origin, number of images, and number of crops per dataset. Table I summarizes the origin, number of images, and number of crops per dataset for the six source datasets, that together form the MLD dataset.

\begin{table}[ht]
    \centering
    \caption{MLD dataset: a summary of the constituent source datasets.}
    \label{tab:datasets}
    \begin{tabular}{l l r r }
        \toprule
        \textbf{Dataset} & \textbf{Origin} & \textbf{Samples} & \textbf{Crops} \\
        \midrule
        PlantVillage~\cite{mohanty2016using}          & USA       & 54{,}306 & 14  \\
        CCMT~\cite{mensah2023ccmt}                     & Ghana     & 24{,}881 & 4 \\
        MAK~\cite{babirye_makerere_maize_2022}        & Uganda    & 46{,}156 & 3 \\
        NMAIST~\cite{mduma_nmaist_bananas_2022}            & Tanzania  & 34{,}345 & 2  \\
        PSFD-Musa~\cite{medhi2022psfd}                 & India     & 5{,}170  & 1 \\ 
        PlantDoc~\cite{singh2020plantdoc}             & India     & 2{,}569  & 13\\\hline
        MLD dataset               & Multi-country   & 167{,}427  & 18  \\
        \bottomrule
    \end{tabular}
\end{table}

These datasets differ in geographical origin, camera devices, backgrounds, and acquisition protocols, allowing us to systematically probe both in-domain and cross-region generalization across different datasets. 

\paragraph{Label curation}
We construct a single unified label space from the six source datasets, organized as a crop–disease hierarchy. Each image is annotated with both a crop label and a disease label, including a healthy class. The full benchmark spans 18 crops and 56 disease classes inclusive of one healthy class per crop. Table \ref{tab:data-count} shows the crops that appear in each dataset.

\begin{table}[htbp]
\centering
\caption{Crop coverage across datasets.}
\label{tab:data-count}
\footnotesize
   \setlength{\tabcolsep}{2.5pt}
\resizebox{\columnwidth}{!}{
\begin{tabular}{l: l l r r r r :r}
\toprule
\textbf{\makecell{Crop \\ (classes)}} & \textbf{\makecell{Plant\\Village}} & \textbf{\makecell{Plant\\Doc}} & \textbf{CCMT} & \textbf{MAK}  & \textbf{NMAIST} & \textbf{\makecell{PSFD\\Musa}} &
\textbf{\makecell{MLD}} \\
\midrule
Apple(4)     & \checkmark & \checkmark &             &  &   & & \checkmark \\
Bellpepper(2)         & \checkmark & \checkmark &   &   &    & & \checkmark       \\
Blueberry(1)       & \checkmark & \checkmark &             & & & & \checkmark         \\
Cherry(2)        & \checkmark & \checkmark &             &   & & & \checkmark         \\
Corn(6)        & \checkmark & \checkmark & \checkmark         & \checkmark & \checkmark  & & \checkmark \\
Grape(4)        & \checkmark & \checkmark &             &      & &   & \checkmark    \\
Orange(1)        & \checkmark &  &             &           & & & \checkmark \\
Peach(2)        & \checkmark & \checkmark &             &    & &  & \checkmark  \\
Potato(3)         & \checkmark & \checkmark &             &     & &   & \checkmark   \\
Soybean(1)        & \checkmark & \checkmark &             &     & & & \checkmark       \\
Raspberry(1)       & \checkmark & \checkmark &             &    & & & \checkmark         \\
Strawberry(2)    & \checkmark & \checkmark &             &      & &   & \checkmark  \\
Squash(1)    & \checkmark & \checkmark &             & & & & \checkmark \\
Tomato(9)        & \checkmark & \checkmark           & \checkmark            &  & &   & \checkmark        \\
Cassava(5)       &            &            & \checkmark  & \checkmark &  & & \checkmark     \\
Beans(3)          &            &            &   &  \checkmark  & & & \checkmark       \\
Cashew(3)       &            &            & \checkmark  &    & & & \checkmark        \\
Bananas(6)     &            &            &  &    &  \checkmark & \checkmark & \checkmark \\
\bottomrule
\end{tabular}}
\end{table}

\paragraph{Data splits}
For each of the six source datasets independently, we apply a stratified split of 80\% for the train set and 20\% for the test set, stratified at the (crop, disease) level so that every joint class is represented in both partitions in proportion to its frequency. The 20\% partitions are reserved as fixed, held-out regional test sets; PlantVillage-test, PlantDoc-test, CCMT-test, MAK-test, NMAIST-test, and PSFD-Musa-test and are never seen during training under any setup. The 80\% partitions are merged to form the training pool. We run stratified 5-fold cross-validation and the resulting model is then evaluated on every held-out regional test set. All results table entries report mean and standard deviation across the 5-fold models.

The combined benchmark contains 167,427 images across multiple crops with diverse disease classes.
By unifying these heterogeneous sources into a single hierarchical label space, we can analyze how a multi-source and multi-crop disease detection model behaves under realistic cross-region domain shifts, rather than only in idealized in-domain settings.

\section{Benchmark Evaluation}

\subsection{Experimental setup}
The experimental design allows us to evaluate different complementary settings for in-region and cross-region performance, single-source and multi-source training of models. 
\begin{enumerate}
    \item \textbf{In-region, single-source (within-domain) evaluation.} A model is trained on a single source dataset. The goal in this setup is to assess how well the model trained on a single dataset generalizes to unseen but in-domain data. PlantVillage and PlantDoc datasets have a relatively wide coverage of crops and their associated diseases, which gives us a better baseline to train a model and test its generalization across multiple crops and diseases. 

 \item \textbf{Cross-region generalization (single-source to unseen regions).} 
A model is trained on i) PlantVillage dataset, which has images with a uniform background taken from a lab setting with a uniform acquisition protocol. The model is thereafter tested and evaluated on a held-out test set from the PlantVillage dataset and on an unseen region of PlantDoc. 

ii) The model is trained on the PlantDoc dataset, whose images have cluttered leaves and a non-uniform acquisition protocol. The model is thereafter tested and evaluated on a held-out test set of the PlantDoc dataset and on an unseen test set from PlantVillage. 

\item   \textbf{Unified multi-source learning with region-wise evaluation.}
The model is trained on a pooled dataset (MLD) from PlantVillage, PlantDoc, CCMT, MAK, NMAIST, and PSFD-Musa, then evaluated on held-out test sets for each of those six datasets.
The goal in this setup is to examine the effect of combining datasets from different regions into a single training set and thereafter evaluate performance on different unseen regions. 

\end{enumerate}

\subsection{Validation metrics and cross-validation protocol}
For each model, we report Crop F1 and Disease F1, both computed as macro-averages across all (crop, disease) classes to account for class imbalance. Disease F1 is reported in the predicted-crop setting, in which the disease head is selected according to the crop prediction of the model. This reflects a potential realistic deployment, where the true crop is not known at inference time. 

\subsection{Baselines}
We compare two approaches for leaf-disease recognition. The first is a \textbf{baseline flat classifier} which treats all crop-disease combinations ($N$) as independent classes in a single multi-class prediction task. Given a feature vector $z$, 
a linear classifier $W_N \in \mathbb{R}^{N \times d}$ produces the predicted class: 
$\hat{y} = \arg\max_k\,(\operatorname{softmax}(W_Nz))(k)$.\\

The second is a \textbf{hierarchical classifier (HiLeaD)}, which first predicts the type of crop and then applies a crop-specific disease head. This enforces a biologically valid prediction and enables evaluation at both crop and disease level. Given the same feature vector $z$, prediction proceeds in two stages.
\paragraph{Stage 1: Crop Prediction}
A linear classifier $W_c \in \mathbb{R}^{C \times d}$, where C is the number of crop categories, produces the crop probabilities 
$\hat{y}_c = \operatorname{softmax}(W_cz)$ and predicted crop
$\hat{c} = \arg\max_k\,(\hat{y}_c)(k)$

\paragraph{Stage 2: Disease Prediction}
Each crop $c$ has a dedicated head $W^{(d)}_c \in \mathbb{R}^{D_c \times d}$, where $D_c$ is the number of diseases for that crop. During training, the disease loss is computed only for the head corresponding to the ground-truth crop $c^*$. During inference, $\hat{c}$ selects the active head, yielding the final prediction pair ($\hat{c}$, $\hat{d}$) where $\hat{d} = \arg\max_k\Big(\operatorname{softmax}\big(W^{(d)}_{\hat{c}}z\big)\Big)(k)$.\\

Given a training image with ground-truth crop $c^*$ and ground-truth disease $d^*$, the total loss combines the crop and disease level cross-entropy, where the first part is $\mathcal{L}_{\text{crop}}$ and the second part is $\mathcal{L}_{\text{disease}}$:

\begin{equation}
\mathcal{L}_{tot} = \underbrace{-\log \hat{y}_c(c^*)}_{\mathcal{L}_{\text{crop}}} + \underbrace{-\log \left( \mathrm{softmax}\!\left( W_{c^*}^{(d^*)} z \right) \right)}_{\mathcal{L}_{\text{disease}}}
\end{equation}

Both the flat and hierarchical classifiers are trained under the same conditions, ensuring a fair comparison between the two formulations.

\subsection{Implementation details}
Both models use a ResNet-18~\cite{resnet} backbone pretrained on ImageNet, with the first three stages frozen and the final two stages and classification head fine-tuned, producing the same feature vector $z$. Training uses a batch size of 32 and early stopping based on unweighted validation loss with patience of 10 (ten) epochs. To address class imbalance, a \texttt{WeightedRandomSampler} and inverse-frequency loss weighting are applied during training. Validation loss is kept unweighted to provide an imbalance-agnostic stopping signal. Models are optimized with Adam (weight decay 1e-4) using parameter-group learning rates and a ReduceLROnPlateau schedule. Training augmentations include random 224×224 cropping from 256×256, horizontal/vertical flips, ±15° rotation, color jitter, and random grayscale. All experiments use random seed 42 and five-fold stratified split.

\section{Results and Discussion}
\label{sec:blind}
We organize the results around the three evaluation protocols defined in Section III. Table III reports macro-F1 for crop and disease classification across all regions and training setups. We report results for both the flat classifier and HiLeaD. 
All numbers are presented as mean$_{\text{std}}$ F1-score. 

\begin{table}[ht]
\footnotesize
\caption{Region-wise cross-evaluation performance under different training setups. Results are noted using mean$_{\text{std}}$ F1-Score.}
\label{tab:combined_results}

\setlength{\tabcolsep}{2pt}

\begin{subtable}{\columnwidth}
\centering
\caption{Model trained on PlantVillage.}
\label{tab:plantvillage}
\begin{tabular}{lcccc}
\toprule
& \multicolumn{2}{c}{Trained on PlantVillage (Flat)} & \multicolumn{2}{c}{Trained on PlantVillage (HiLeaD)} \\
\cmidrule(r){2-3} \cmidrule(l){4-5}
Region & Crop F1 & Disease F1 & Crop F1 & Disease F1 \\
\midrule
PlantVillage  & 99.39$_{\text{0.11}}$ & 98.91$_{\text{0.24}}$ & \textbf{99.67$_{\text{0.11}}$}  & \textbf{99.07$_{\text{0.20}}$}  \\
PlantDoc      & 23.29$_{\text{1.36}}$ & 11.46$_{\text{1.10}}$ & \textbf{24.11$_{\text{1.38}}$}  & \textbf{12.88$_{\text{1.49}}$}  \\
\bottomrule
\end{tabular}
\end{subtable}

\vspace{0.6em}

\begin{subtable}{\columnwidth}
\centering
\caption{Model trained on PlantDoc.}
\label{tab:plantdoc}
\begin{tabular}{lcccc}
\toprule
& \multicolumn{2}{c}{Trained on PlantDoc (Flat)} & \multicolumn{2}{c}{Trained on PlantDoc (HiLeaD)} \\
\cmidrule(r){2-3} \cmidrule(l){4-5}
Region & Crop F1 & Disease F1 & Crop F1 & Disease F1 \\
\midrule
PlantVillage   & 40.93$_{\text{1.74}}$ & 31.87$_{\text{1.23}}$ & \textbf{43.05$_{\text{3.24}}$}  & \textbf{34.08$_{\text{2.15}}$}  \\
PlantDoc       & 73.91$_{\text{1.64}}$ & 57.37$_{\text{1.67}}$ & \textbf{75.64$_{\text{1.81}}$}  & \textbf{58.09$_{\text{2.38}}$}  \\
\bottomrule
\end{tabular}
\end{subtable}

\vspace{0.6em}

\begin{subtable}{\columnwidth}
\centering
\caption{Model trained on the unified pooled multi-crop dataset (MLD).}
\label{tab:mld_results}
\begin{tabular}{lcccc}
\toprule
& \multicolumn{2}{c}{Trained on MLD (Flat)} & \multicolumn{2}{c}{Trained on MLD (HiLeaD)} \\
\cmidrule(r){2-3} \cmidrule(l){4-5}
Region & Crop F1 & Disease F1 & Crop F1 & Disease F1 \\
\midrule
PlantVillage & 94.07$_{\text{4.61}}$ & 96.60$_{\text{1.31}}$ & \textbf{99.63$_{\text{0.09}}$}  & \textbf{98.87$_{\text{0.23}}$} \\
PlantDoc     & 56.58$_{\text{1.92}}$ & 37.93$_{\text{1.29}}$ & \textbf{60.51$_{\text{1.46}}$}  & \textbf{39.64$_{\text{1.63}}$}  \\
CCMT         & 57.87$_{\text{7.48}}$ & 53.33$_{\text{4.94}}$ & \textbf{59.12$_{\text{12.16}}$}  & \textbf{55.87$_{\text{3.80}}$}  \\
MAK          & 70.44$_{\text{18.87}}$ & 50.49$_{\text{15.48}}$ & \textbf{66.89$_{\text{10.28}}$}  & \textbf{55.37$_{\text{11.50}}$}  \\
NMAIST       & 99.68$_{\text{0.13}}$ & 58.55$_{\text{9.28}}$ & \textbf{99.82$_{\text{0.12}}$} & \textbf{63.09$_{\text{9.71}}$}  \\
PSFD-Musa    & 89.97$_{\text{20.06}}$ & 95.77$_{\text{8.19}}$ & \textbf{100.00$_{\text{0.00}}$} & \textbf{99.58$_{\text{0.21}}$}  \\
\bottomrule
\end{tabular}
\end{subtable}

\end{table}

\paragraph {Within‑region single source evaluation}
When the model is trained and tested on PlantVillage, seen in Table \ref{tab:plantvillage}, both formulations achieve near-perfect performance of 98.91\% disease F1 for the flat classifier and 99.07\% disease F1 for HiLeaD. This is the regime in which prior work typically reports results, and it produces accuracies consistent with the 84.0–99.8\% range commonly seen in multi-crop disease recognition studies. PlantVillage is captured under uniform laboratory conditions with a consistent acquisition protocol, and the in-domain test split inherits these properties, so a model can fit the data closely.

The PlantDoc-trained model displays a notably inconsistent intra-domain behaviour as shown in Table \ref{tab:plantdoc}. Its in-domain disease F1 scores are 57.37\% (flat) and 58.09\% (HiLeaD), despite comparable label complexity to PlantVillage. PlantDoc is an order of magnitude smaller (Table \ref{tab:datasets}) and consists of in-the-wild field images with cluttered backgrounds, inconsistent lighting, and variable acquisition protocols. 
 
This contrast already exposes a core problem this benchmark is designed to study: in-domain F1 score is strongly influenced by acquisition conditions rather than the recognition difficulty.

\paragraph{Cross-region single-source generalization}
Cross-region results in Table \ref{tab:plantvillage} and \ref{tab:plantdoc}, reveal the severity of domain shift. A model trained on PlantVillage and evaluated on PlantDoc collapses from 98.91\% to 11.46\% disease F1 (flat), an 87 percentage point drop. HiLeaD shows a similar pattern, dropping from 99.07\% to 12.88\%. The architecture alone cannot bridge this gap. Both formulations fail similarly when the test distribution differs sharply from the training distribution. This finding is consistent with the broader observation in studies that, single-dataset evaluation overstates model performance. 
On the other hand, a PlantDoc-trained model evaluated on PlantVillage reaches 31.87\% (flat) and 34.08\% (HiLeaD), far from in-domain numbers but nearly three times higher than the PlantVillage-to-PlantDoc transfer. This asymmetry suggests that exposure to noisy, cluttered field images during training yields representations that partially transfer to clean laboratory conditions, while the reverse does not hold.

\begin{table}[ht]
\centering
\caption{Per-crop performance (mean$_{\text{std}}$ F1-score) for crop classification and disease recognition across datasets on MLD trained HiLeaD }
\label{tab:perclass_f1_stacked}
\scriptsize
\setlength{\tabcolsep}{1.5pt}

\begin{tabular}{lcccccc}
\toprule
& \textbf{PlantVillage} & \textbf{PlantDoc} & \textbf{CCMT} & \textbf{MAK} & \textbf{NMAIST} & \textbf{PSFD-Musa} \\
\cmidrule(lr){2-2} \cmidrule(lr){3-3} \cmidrule(lr){4-4} \cmidrule(lr){5-5} \cmidrule(lr){6-6} \cmidrule(lr){7-7}
\textbf{Crop} 
& \begin{tabular}[c]{@{}c@{}}Crop F1\\Disease F1\end{tabular}
& \begin{tabular}[c]{@{}c@{}}Crop F1\\Disease F1\end{tabular}
& \begin{tabular}[c]{@{}c@{}}Crop F1\\Disease F1\end{tabular}
& \begin{tabular}[c]{@{}c@{}}Crop F1\\Disease F1\end{tabular}
& \begin{tabular}[c]{@{}c@{}}Crop F1\\Disease F1\end{tabular}
& \begin{tabular}[c]{@{}c@{}}Crop F1\\Disease F1\end{tabular} \\
\midrule

Apple        & \begin{tabular}{c}99.76 $_{\text{0.05}}$\\99.73 $_{\text{2.42}}$\end{tabular}
             & \begin{tabular}{c}76.64 $_{\text{1.15}}$\\52.12 $_{\text{0.12}}$\end{tabular}
             & -- & -- & -- & --\\ \hdashline

Banana       & -- & -- & -- & -- 
            & \begin{tabular}{c}99.83 $_{\text{0.08}}$\\49.38 $_{\text{0.19}}$\end{tabular}
             & \begin{tabular}{c}100.00 $_{\text{0.00}}$\\66.38 $_{\text{0.14}}$\end{tabular} \\ \hdashline

Bellpepper  & \begin{tabular}{c}99.72 $_{\text{0.04}}$\\99.66 $_{\text{0.22}}$\end{tabular}
             & \begin{tabular}{c}60.03 $_{\text{3.70}}$\\67.95 $_{\text{5.52}}$\end{tabular}
             & -- & -- & -- & --\\ \hdashline

Cherry       & \begin{tabular}{c}99.90 $_{\text{0.05}}$\\99.92 $_{\text{0.11}}$\end{tabular}
             & \begin{tabular}{c}49.72 $_{\text{8.32}}$\\33.60 $_{\text{3.73}}$\end{tabular}
             & -- & -- & -- & -- \\ \hdashline
             
Blueberry    & \begin{tabular}{c}98.37 $_{\text{0.89}}$\\100 $_{\text{0.00}}$ \end{tabular}
             & \begin{tabular}{c}65.12 $_{\text{4.63}}$\\83.89 $_{\text{3.07}}$\end{tabular}
             & -- & -- & -- & -- \\ \hdashline

Beans       & -- & -- & --
            & \begin{tabular}{c}99.89 $_{\text{0.05}}$\\92.36 $_{\text{0.24}}$\end{tabular}
            & -- & -- \\ \hdashline
            
Corn         & \begin{tabular}{c}99.99 $_{\text{0.03}}$\\64.60 $_{\text{0.22}}$\end{tabular}
             & \begin{tabular}{c}97.55 $_{\text{0.57}}$\\36.30 $_{\text{2.26}}$\end{tabular}
             & \begin{tabular}{c}99.81 $_{\text{0.11}}$\\55.16 $_{\text{0.77}}$\end{tabular}
             & \begin{tabular}{c}99.94 $_{\text{0.05}}$\\49.03 $_{\text{0.24}}$\end{tabular}
             & \begin{tabular}{c}99.79 $_{\text{0.10}}$\\48.75 $_{\text{0.20}}$\end{tabular}
             & -- \\ \hdashline

Grape        & \begin{tabular}{c}99.99 $_{\text{0.02}}$\\99.72 $_{\text{0.11}}$\end{tabular}
             & \begin{tabular}{c}77.52 $_{\text{3.11}}$\\37.38 $_{\text{1.25}}$\end{tabular}
             & -- & -- & -- & -- \\ \hdashline

Peach        & \begin{tabular}{c}99.89 $_{\text{0.07}}$\\99.44 $_{\text{0.40}}$\end{tabular}
             & \begin{tabular}{c}76.66 $_{\text{4.37}}$\\42.95 $_{\text{2.77}}$\end{tabular}
             & -- & -- & -- & -- \\ \hdashline

Orange      & \begin{tabular}{c}99.94 $_{\text{0.04}}$\\99.96 $_{\text{0.03}}$\end{tabular}
             & -- & -- & -- & -- & --\\ \hdashline

Potato       & \begin{tabular}{c}98.29 $_{\text{0.73}}$\\98.36 $_{\text{0.99}}$\end{tabular}
             & \begin{tabular}{c}52.95 $_{\text{6.14}}$\\33.15 $_{\text{8.39}}$\end{tabular}
             & -- & -- & -- & --  \\ \hdashline

Strawberry   & \begin{tabular}{c}99.97 $_{\text{0.06}}$\\100.0 $_{\text{0.00}}$\end{tabular}
             & \begin{tabular}{c}87.82 $_{\text{4.28}}$\\48.04 $_{\text{0.69}}$\end{tabular}
             & -- & -- & -- & -- \\ \hdashline

Tomato       & \begin{tabular}{c}99.72 $_{\text{0.09}}$\\86.10 $_{\text{0.27}}$\end{tabular}
             & \begin{tabular}{c}76.47 $_{\text{2.94}}$\\37.09 $_{\text{1.99}}$\end{tabular}
             & \begin{tabular}{c}99.57 $_{\text{0.13}}$\\41.45 $_{\text{1.24}}$\end{tabular}
             & -- & -- & --\\ \hdashline

Soybean   & \begin{tabular}{c}99.62 $_{\text{0.19}}$\\99.61 $_{\text{0.19}}$\end{tabular}
             & \begin{tabular}{c}61.69 $_{\text{14.54}}$\\63.26 $_{\text{14.8}}$\end{tabular}
             & -- & -- & -- & -- \\ \hdashline

Raspberry   & \begin{tabular}{c}99.86 $_{\text{0.27}}$\\99.86 $_{\text{0.27}}$\end{tabular}
             & \begin{tabular}{c}76.07 $_{\text{4.68}}$\\82.22 $_{\text{3.91}}$\end{tabular}
             & -- & -- & -- & --\\ \hdashline

Squash      & \begin{tabular}{c}99.97 $_{\text{0.05}}$\\99.97 $_{\text{0.05}}$\end{tabular}
             & \begin{tabular}{c}94.35 $_{\text{1.14}}$\\95.17 $_{\text{1.70}}$\end{tabular}
             & -- & -- & -- & --\\ \hdashline
    
Cassava     & -- & -- 
            & \begin{tabular}{c}99.57 $_{\text{0.16}}$\\76.36 $_{\text{0.49}}$\end{tabular}
             & \begin{tabular}{c}99.84 $_{\text{0.10}}$\\56.51 $_{\text{0.64}}$ \end{tabular}
            & -- & -- \\ \hdashline 
            
Cashew      & -- & --
            & \begin{tabular}{c}99.71 $_{\text{0.09}}$\\98.77 $_{\text{0.14}}$\end{tabular}
            & -- & -- & -- 
            \\ 

\bottomrule
\end{tabular}
\end{table}

\paragraph{Pooled multi‑source training and region‑wise generalization}
Training on our introduced MLD benchmark, i.e. combining all six source datasets, improves cross-region disease F1 Scores as seen in Table \ref{tab:mld_results}. For HiLeaD, disease F1 reaches 39.64\% on PlantDoc, 55.87\% on CCMT, 55.37\% on MAK, 63.09\% on NMAIST, and 99.58\% on PSFD-Musa, while preserving in-domain PlantVillage performance at 98.87\%. The improvement on PlantDoc is three times better than single-source training, achieved without architectural changes.

Two patterns emerge from the pooled-training results. First, under pooled training, HiLeaD loses only 0.20 percentage points of in-domain F1, compared to a 2.31 percentage point drop in the flat baseline. This robustness stems from predicting disease within each crop independently, which reduces the confusion introduced by the larger label space. Second, the hierarchical advantage grows under pooled training, with gains ranging from 1.71 percentage points on PlantDoc to 4.88 percentage points on MAK. Crop conditioning appears to be more useful when the label space is heterogeneous and the test distribution far from the training distribution.

Table \ref{tab:perclass_f1_stacked} breaks down the performance of HiLeaD, per crop per region under the MLD trained model. Crop F1 stays above 99\% across PlantVillage, CCMT, MAK, and NMAIST but degrades to a 50-80\% range on PlantDoc. Disease F1 scores follow a similar pattern within PlantVillage: most crops reach 99\%+ disease F1, though a few drop notably (Corn, Cassava, Tomato), and PlantDoc collapses to 33-50\% on nearly all crops. 
Corn, the only crop represented in all regions (except the PSFD-Musa dataset), retains 97-99\% crop F1, but the disease F1 ranges from 36.30\% on PlantDoc to 64.60\% on PlantVillage, confirming that acquisition conditions contribute highly to recognition difficulty. Larger per-crop standard deviations in the field datasets further suggest that fold-to-fold variation is concentrated where the training distribution covers the target region most thinly.

\paragraph{Comparison of the flat and hierarchical classifier baselines} 
Across all setups in Table \ref{tab:combined_results}, two facts hold simultaneously. Pooled multi-source training accounts for most of the recovery from the single-source collapse. HiLeaD provides a small gain, ranging from 0.16 percentage points on PlantVillage (in-domain) to 0.72 percentage points on PlantDoc (in-domain). Under the MLD benchmark, the hierarchical formulation offers a small additional gain of 1.71-4.88 percentage points. Data diversity addresses the distribution shift, while hierarchical conditioning addresses the label-space heterogeneity that data diversity creates.

\section{Conclusion}
\label{sec:concl}
We introduced the MLD dataset, a unified multi-region benchmark for multi-crop leaf disease recognition that combines six public datasets into a shared hierarchical taxonomy spanning 18 crops and 56 disease classes, including one healthy class per crop. We set out to test whether the near-perfect accuracies reported on standard crop-disease benchmarks survive a realistic cross-region domain shift. Our experiments yield three findings. 

First, models trained on a single source dataset exhibit a severe in-domain versus cross-region gap. A PlantVillage-trained HiLeaD model drops from 99.07\% to 12.88\% on the Disease F1 score on PlantDoc, while the reverse transfer reaches 34.08\%, nearly three times higher. This asymmetry suggests that exposure to noisy field images during training yields representations that are more transferable than those from laboratory data alone. 
Second, pooled multi-source training partially closes the cross-region performance gap, raising cross-region disease F1 from 12.88\% to 39.64\% on PlantDoc using the same model architecture. Per-class analysis confirms that crop recognition remains robust, while disease F1 scores vary across regions for the same crop. Corn ranges from 36.30\% on PlantDoc to 64.60\% on PlantVillage, showing that acquisition conditions are a primary driver of recognition difficulty.

Third, the hierarchical formulation (HiLeaD) provides a consistent gain over the flat baseline (1.71 to 4.88 percentage points disease F1), most pronounced on the pooled MLD dataset where label-space heterogeneity is the highest.  

These findings have practical implications for deploying disease recognition models for smallholder farmers, where target conditions differ notably from curated benchmarks. Specifically, laboratory-only training should not be expected to generalize to field conditions, and training-data diversity has a greater impact than model design. We release the unified taxonomy and evaluation splits to support reproducible benchmarking.

\bibliographystyle{IEEEtran}
\bibliography{references}

@article{hlatshwayo2023determinants,
  title={The determinants of crop productivity and its effect on food and nutrition security in rural communities of South Africa},
  author={Hlatshwayo, Simphiwe Innocentia and Ngidi, Mjabuliseni Simon Cloapas and Ojo, Temitope Oluwaseun and Modi, Albert Thembinkosi and Mabhaudhi, Tafadzwanashe and Slotow, Rob},
  journal={Frontiers in sustainable food systems},
  volume={7},
  pages={1091333},
  year={2023},
  publisher={Frontiers Media SA}
}

@misc{resnet,
      title={Deep Residual Learning for Image Recognition}, 
      author={Kaiming He and Xiangyu Zhang and Shaoqing Ren and Jian Sun},
      year={2015},
      eprint={1512.03385},
      archivePrefix={arXiv},
      primaryClass={cs.CV},
      url={https://arxiv.org/abs/1512.03385}, 
}

@article{giller2021small,
  title={Small farms and development in sub-Saharan Africa: Farming for food, for income or for lack of better options?},
  author={Giller, Ken E and Delaune, Thomas and Silva, Jo{\~a}o Vasco and van Wijk, Mark and Hammond, James and Descheemaeker, Katrien and van de Ven, Gerrie and Schut, Antonius GT and Taulya, Godfrey and Chikowo, Regis and others},
  journal={Food Security},
  volume={13},
  number={6},
  pages={1431--1454},
  year={2021},
  publisher={Springer}
}

@article{basantaray2024crop,
  title={Crop diversification and income of agricultural households in India: an empirical analysis},
  author={Basantaray, AK and Acharya, Shailesh and Patra, Tophan},
  journal={Discover Agriculture},
  volume={2},
  number={1},
  pages={8},
  year={2024},
  publisher={Springer}
}

@article{islam2024determinants,
  title={Determinants of crop diversification and its impact on farmers' income: A case study in Rangpur District, Bangladesh},
  author={Islam, Md Sayemul and Jahan, Hasneen and Ema, Nishat Sultana and Ahmed, Md Rubel},
  journal={JSFA Reports},
  volume={4},
  number={10},
  pages={352--361},
  year={2024},
  publisher={Wiley Online Library}
}

@article{mohanty2016using,
  title={Using deep learning for image-based plant disease detection},
  author={Mohanty, Sharada P and Hughes, David P and Salath{\'e}, Marcel},
  journal={Frontiers in plant science},
DOI={10.3389/fpls.2016.01419},
  volume={7},
  pages={215232},
  year={2016},
  publisher={frontiers}
}

@incollection{singh2020plantdoc,
  title={PlantDoc: A dataset for visual plant disease detection},
  author={Singh, Davinder and Jain, Naman and Jain, Pranjali and Kayal, Pratik and Kumawat, Sudhakar and Batra, Nipun},
  booktitle={Proceedings of the 7th ACM IKDD CoDS and 25th COMAD},
  pages={249--253},
  year={2020}
}

@article{mensah2023ccmt,
  title={CCMT: Dataset for crop pest and disease detection},
  author={Mensah, Patrick Kwabena and Akoto-Adjepong, Vivian and Adu, Kwabena and Ayidzoe, Mighty Abra and Bediako, Elvis Asare and Nyarko-Boateng, Owusu and Boateng, Samuel and Donkor, Esther Fobi and Bawah, Faiza Umar and Awarayi, Nicodemus Songose and others},
  journal={Data in Brief},
  volume={49},
  pages={109306},
  year={2023},
  publisher={Elsevier}
}

@article{nalwanga2025multi,
  title={Multi-Crop Disease Detection in Computer Vision for Resource-Constrained Farms—A Review},
  author={Nalwanga, Rosemary and Spreeuwers, Luuk and Talavera, Estefania and Owomugisha, Godliver},
  journal={IEEE Access},
  volume={13},
  pages={217265--217284},
  year={2025},
  publisher={IEEE}
}

@article{moupojou2024segment,
  title={Segment anything model and fully convolutional data description for plant multi-disease detection on field images},
  author={Moupojou, Emmanuel and Retraint, Florent and Tapamo, Hyppolite and Nkenlifack, Marcellin and Kacfah, Cheikh and Tagne, Appolinaire},
  journal={IEEE Access},
  volume={12},
  pages={102592--102605},
  year={2024},
  publisher={IEEE}
}

@inproceedings{pathak2025cnns,
  title={CNNs, Transformers, and Beyond: AI Models for Crop Disease Detection-A Review},
  author={Pathak, Deepak Kumar and Goel, Lavika and Mishra, Aishwarya},
  booktitle={2025 International Conference on Emerging Technologies in Electronics and Green Energy (ICETEG)},
  pages={1--6},
  year={2025},
  organization={IEEE}
}

@article{sharma2023dlmc,
  title={DLMC-Net: Deeper lightweight multi-class classification model for plant leaf disease detection},
  author={Sharma, Vivek and Tripathi, Ashish Kumar and Mittal, Himanshu},
  journal={Ecological informatics},
  volume={75},
  pages={102025},
  year={2023},
  publisher={Elsevier}
}

@article{kundu2022disease,
  title={Disease detection, severity prediction, and crop loss estimation in MaizeCrop using deep learning},
  author={Kundu, Nidhi and Rani, Geeta and Dhaka, Vijaypal Singh and Gupta, Kalpit and Nayaka, Siddaiah Chandra and Vocaturo, Eugenio and Zumpano, Ester},
  journal={Artificial intelligence in agriculture},
  volume={6},
  pages={276--291},
  year={2022},
  publisher={Elsevier}
}

@article{hamid2025enhancing,
  title={Enhancing agricultural disease detection: a multi-model deep learning novel approach},
  author={Hamid, Muhammad Khalid and Shah, Said Khalid and Husnain, Ghassan and Ghadi, Yazeed Yasin and Al Maaytah, Shahab Ahmad and Qahmash, Ayman},
  journal={Engineering Reports},
  volume={7},
  number={1},
  pages={e13113},
  year={2025},
  publisher={Wiley Online Library}
}

@inproceedings{saeed2021multi,
  title={A multi-crop disease detection and classification approach using cnn},
  author={Saeed, Zubair and Raza, Ali and Qureshi, Ans H and Yousaf, Muhammad Haroon},
  booktitle={2021 International Conference on Robotics and Automation in Industry (ICRAI)},
  pages={1--6},
  year={2021},
  organization={IEEE}
}

@article{saleem2024comparison,
  title={Comparison of deep learning models for multi-crop leaf disease detection with enhanced vegetative feature isolation and definition of a new hybrid architecture},
  author={Saleem, Sajjad and Sharif, Muhammad Irfan and Sharif, Muhammad Imran and Sajid, Muhammad Zaheer and Marinello, Francesco},
  journal={Agronomy},
  volume={14},
  number={10},
  pages={2230},
  year={2024},
  publisher={MDPI}
}

@article{medhi2022psfd,
  title={PSFD-Musa: A dataset of banana plant, stem, fruit, leaf, and disease},
  author={Medhi, Epsita and Deb, Nabamita},
  journal={Data in brief},
  volume={43},
  pages={108427},
  year={2022},
  publisher={Elsevier}
}

@misc{babirye_makerere_maize_2022,
author = {Claire Babirye and Joyce Nakatumba-Nabende and Gloria Namanya and Chodrine Mutebi and Moses Ebellu and Joab Murungi and Saolo Tobius and Jonah Ssemwogerere and Annet Nakayima and Deborah Nabagereka and Judith Asasira and Ruth Kanyesigye},
title = {Makerere University Maize Image Dataset},
year = {2022},
publisher = {Harvard Dataverse},
doi = {10.7910/DVN/LPGHKK},
url = {https://doi.org/10.7910/DVN/LPGHKK},
}

@misc{mduma_nmaist_bananas_2022,
author = {Neema Mduma and Hudson Laizer and Loyani Loyani and Mbwana Macheli and Zablon Msengi and Alice Karama and Irine Msaki and Sophia Sanga and Kennedy Jomanga and Leo Judith},
title = {The Nelson Mandela African Institution of Science and Technology Bananas Dataset},
year = {2022},
publisher = {Harvard Dataverse},
doi = {10.7910/DVN/LQUWXW},
url = {https://doi.org/10.7910/DVN/LQUWXW}
}

@inproceedings{shalini2025machine,
  title={Machine learning based leaf disease diagnosis system},
  author={Shalini, V Baby and Kumar, Bittu and Varma, P Varun and Reddy, P Vinay Kumar and Kumar, T Tarun},
  booktitle={2025 International Conference on Multi-Agent Systems for Collaborative Intelligence (ICMSCI)},
  pages={1124--1128},
  year={2025},
  organization={IEEE}
}

@article{gao2025symptom,
  title={Symptom development in plant viral diseases: what, how, and why?},
  author={Gao, Man and Lozano-Dur{\'a}n, Rosa},
  journal={Annual Review of Phytopathology},
  volume={63},
  year={2025},
  publisher={Annual Reviews}
}

@article{dilliwar2025plant,
  title={Plant Disease Detection Using Convolutional Neural Networks (CNNs)},
  author={Dilliwar, Kuldeep and Mandal, Subham and Sahu, Nagesh and Shriwas, Manish Kumar and Lanjewar, Rajeshri},
  journal={IJSAT-International Journal on Science and Technology},
  volume={16},
  number={2},
  year={2025},
  publisher={International Research Publication and Journals}
}

\end{document}